\documentclass[runningheads,]{llncs}
\usepackage[position=above,caption=false]{subfig}
\usepackage{xcolor}
\usepackage{lipsum}
\usepackage{graphicx}
\usepackage[export]{adjustbox}
\usepackage{makecell}
\usepackage{multicol}
\usepackage{booktabs}
\usepackage{lscape} 
\usepackage{hyperref}
\title{Transfer Learning between Motor Imagery Datasets using Deep Learning - Validation of Framework and Comparison of Datasets
\thanks{Supported by the Donders Center for Cognition (DCC).}}
\titlerunning{Transfer Learning between Motor Imagery datasets using Deep Learning}
\author{Pierre~Guetschel\inst{1,2}
\orcidID{0000-0002-8933-7640} \and
Michael~Tangermann\inst{1,2}
\orcidID{0000-0001-6729-0290}}
\authorrunning{P. Guetschel \and M. Tangermann}
\institute{Donders Institute for Brain, Cognition and Behaviour,\\
Radboud University, Nijmegen, Netherlands \and
\email{<first>.<last>@donders.ru.nl}
}

\begin{document}
\maketitle              
\begin{abstract}

We present a simple deep learning-based framework commonly used in computer vision and demonstrate its effectiveness for cross-dataset transfer learning in mental imagery decoding tasks that are common in the field of Brain-Computer Interfaces (BCI). We investigate, on a large selection of 12 motor-imagery datasets, which ones are well suited for transfer, both as donors and as receivers.
%
\\
\textit{Challenges.}
Deep learning models typically require long training times and are data-hungry, which impedes their use for BCI systems that have to minimize the recording time for (training) examples and are subject to constraints induced by experiments involving human subjects. A solution to both issues is transfer learning, but it comes with its own challenge, i.e., substantial data distribution shifts between datasets, subjects and even between subsequent sessions of the same subject.
\\ 
\textit{Approach.}
For every pair of pre-training (donor) and test (receiver) dataset, we first train a model on the donor before training merely an additional new linear classification layer based on a few receiver trials. Performance of this transfer approach is then tested on other trials of the receiver dataset.
%
\\
\textit{Significance.}
First, we lower the threshold to use transfer learning between motor imagery datasets: the overall framework is extremely simple and nevertheless obtains decent classification scores.  Second, we demonstrate that deep learning models are a good option for motor imagery cross-dataset transfer both for the reasons outlined in the first point and because the framework presented is viable in online scenarios. 
Finally, analysing which datasets are best suited for transfer learning can be used as a reference for future researchers to determine which to use for pre-training or benchmarking.


\end{abstract}
\pagebreak


\section{Introduction} \label{introduction}
Electroencephalography (EEG) offers a non-invasive modality for capturing the electrical activity of the brain. Utilizing machine learning techniques, specific brain activities can be decoded in single trial.
For instance, event-related potentials (ERPs) signify the brain's response to the occurrence of an awaited and attended event~\cite{luckIntroductionEventRelatedPotential2014}. 
Likewise, the sensorimotor cortices, being relatively large areas of the cortical surface, allow for the decoding of motor execution and motor imagery activities. The decoded brain activities can be mapped to computer commands. In so-called Brain-Computer Interfaces (BCIs)~\cite{clercBrainComputerInterfaces2016}, the commands are used to drive applications that are directly controlled by the brain.

In a typical setting, each BCI session starts with an initial phase where the user is prompted to produce predefined outputs. This is typically called the \textit{calibration phase}. The acquired data enables the supervised training of a decoding model. 
Only afterwards, the user can interact freely with the BCI application. This productive phase is typically called the \textit{online mode}. 
However, calibration can be a tedious and uninteresting process for a user, and may require up to an hour of attention-demanding activities, even though the interesting phase we aim to start with as quickly as possible is the online phase.

Specifically for BCIs that make use of motor imagery tasks, the tedious calibration process prompts the consideration of Transfer Learning (TL) methods. TL leverages information from various sources, such as earlier sessions of the same user, of other users, or datasets obtained under slightly different experimental conditions, to reduce the need for calibration data samples collected during the ongoing session. In the most rudimentary approach to TL, a model is trained on a dataset we will call \textit{donor}, and tested on another one we will call \textit{receiver}. The necessity for TL becomes even more pronounced in the context of deep learning, given its inherent requirements for large training datasets and a substantial training time which would further delay the online phase~\cite{iangoodfellowDeepLearning2016}. However, TL presents its own challenges in the form of distributional shifts across different datasets, subjects, or even sessions. They can be attributed to factors like variations in experimental protocol, EEG recording systems, brain morphology, level of fatigue, medication intake, and more~\cite{jayaramTransferLearningBraincomputer2016}.
While TL has been widely studied in BCI, the focus has been mostly on cross-session transfer~\cite{zhuEEGNetEnsembleLearning2021}, cross-subject transfer~\cite{samekTransferringSubspacesSubjects2013,jeonMutualInformationDrivenSubjectInvariant2021}, or both~\cite{koblerSPDDomainspecificBatch2022}.
However, cross-dataset transfer learning has been a relatively underexplored domain. The recent BEETL competition has begun to fill this void by drawing attention to the challenges and opportunities cross-dataset transfer represents for BCI~\cite{wei2021BEETLCompetition2022}. Notably, the top three winning methods in this competition were based on deep learning, thus indicating promising potential for such models.
Furthermore, it was recently suggested to use cross-dataset transfer to evaluate the structure of cognitive tasks~\cite{aristimunhaEvaluatingStructureCognitive2023}.

In this study, we extend a TL method already standard in the field of computer vision, the ImageNet linear evaluation protocol~\cite{dengImageNetLargescaleHierarchical2009}, to BCIs. Following our previous work which showcased its efficacy for cross-subject transfer~\cite{GuePapTan22}, here we focus on its applicability for cross-dataset transfer. We experiment with one-to-one transfer across twelve motor imagery datasets, leveraging the MOABB library~\cite{jayaramMOABBTrustworthyAlgorithm2018} for both, dataset acquisition and normalized evaluation.

In this article, we will examine several pertinent questions: 
How challenging is actually TL between motor imagery datasets? 
Is deep learning genuinely a good option for TL in BCI? 
If so, what constitutes the minimal configuration necessary?
Which datasets serve as the most effective donors?
And which datasets are the best receivers?
Our contributions can be summarized as follows:%
\begin{enumerate}
    \item We present a TL framework that can deal with a minimum of data from the receiver dataset and still reach good performance;
    \item We lower the threshold to use TL between motor imagery datasets considering the extreme simplicity of this framework; 
    \item We provide evidence that deep learning models and our TL framework are apt choices for cross-dataset transfer, further substantiated by the viability of our framework for online scenarios;
    \item We characterize the motor imagery datasets currently available in MOABB;
    \item We present an analytical investigation into the suitability of these datasets for TL, serving as a valuable resource for future research in pre-training or benchmarking models.
\end{enumerate}

Supplementary materials, including source code, pre-trained models, and comprehensive results, are available online\footnote{Source code: \url{https://gitlab.com/PierreGtch/motor_embedding_benchmark}}\footnote{Pre-trained models, and results: \url{https://huggingface.co/PierreGtch/EEGNetv4}}.

\section{Materials} \label{materials}
\subsection{Neural Network Architecture: EEGNet}

We used EEGNet~\cite{lawhernEEGNetCompactConvolutional2018} as neural network architecture to conduct our experiments.
It consists of an initial temporal convolutional layer that captures temporal correlations in EEG signals. This is followed by a depthwise spatial convolution layer that accounts for spatial correlations across electrodes. The network concludes with a separable convolution for dimensionality reduction before reaching the linear classification layer.

Our choice of EEGNet as the neural network architecture is motivated by three primary factors. First, its versatility is demonstrated through proficient performance across multiple BCI paradigms~\cite{lawhernEEGNetCompactConvolutional2018}. Second, its simplicity and computational efficiency make it particularly lightweight. Third, EEGNet has earned widespread community acceptance for EEG decoding, as evidenced by several studies~\cite{xuTransferLearningFramework2021,razaSingleTrialEEGClassification2020,zhuEEGNetEnsembleLearning2021,schneiderQEEGNetEnergyEfficient8bit2020,riyadIncepEEGNetConvNetMotor2020,dengAdvancedTSGLEEGNetMotor2021}.

\subsection{Data: Motor Imagery}

We conducted our experiments on EEG-based motor imagery (MI) datasets. They are constituted of EEG recordings capturing subjects as they either imagine or execute specific motor tasks. These motor tasks can range from imagined actions such as squeezing a fist to wiggling toes.
In each dataset, multiple motor imagery tasks are represented and the classification problem is to determine which task was imagined/executed in each example, which typically consists of a few seconds of EEG recordings called a trial. 

\subsubsection{Description of Datasets}\label{datasets}
\begin{table}[ht]
\centering
\caption{Summary of Datasets Evaluated in the Study.}
\label{table:datasets}
\begin{tabular}{llcccc}
\toprule
Dataset & Classes & \makecell{No.\\classes} & \makecell{No.\\subjects} & \makecell{No.\\sessions} & \makecell{No.\\examples} \\
\midrule
AlexMI~\cite{barachantCommandeRobusteEffecteur2012} & \textit{f, r, rh.} & 3 & 8 & 1 &20 \\ 
BNCI2014001~\cite{tangermannReviewBCICompetition2012}& \textit{f, lh, rh, t.} & 4 & 9 & 2 & 72 \\ 
BNCI2014004~\cite{leebBrainComputerCommunication2007} & \textit{lh, rh.} & 2 & 9 & 5 & 72 \\ 
BNCI2015001~\cite{fallerAutocalibrationRecurrentAdaptation2012} & \textit{f, rh.} & 2 & 13 & \makecell[t]{3 (subj. 8-11),\\2 (others)} & 100 \\ 
BNCI2015004~\cite{schererIndividuallyAdaptedImagery2015} & \textit{f, n, rh, s, wa.} & 5 & 9 & 2 & 39 \\ 
Cho2017~\cite{choEEGDatasetsMotor2017} & \textit{lh, rh.} & 2 & 53 & 1 & 101 \\ 
Lee2019\_MI~\cite{leebBrainComputerCommunication2007} & \textit{lh, rh.} & 2 & 55 & 2 & 200 \\ 
Ofner2017~\cite{ofnerUpperLimbMovements2017} & \makecell[tl]{\itshape r, ree, ref, rhc,\\rho, rp, rs.} & 7 & 15 & 1 & 60 \\ 
PhysionetMI~\cite{goldbergerPhysioBankPhysioToolkitPhysioNet2000a} & \textit{bh, f, lh, r, rh.} & 5 & 109 & 1 & 23 \\ 
Schirrmeister2017~\cite{schirrmeisterDeepLearningConvolutional2017} & \textit{f, lh, r, rh.} & 4 & 14 & 1 & 241 \\ 
Weibo2014~\cite{yiEvaluationEEGOscillatory2014} & \makecell[tl]{\itshape bh, f, lh, lhrf,\\r, rh, rhlf.} & 7 & 10 & 1 & 79 \\ 
Zhou2016~\cite{zhouFullyAutomatedTrial2016} & \textit{f, lh, rh.} & 3 & 4 & 3 & 50 \\ 
\bottomrule
\end{tabular}
\end{table}

In this study, we included all the Motor Imagery datasets available with the MOABB library~\cite{jayaramMOABBTrustworthyAlgorithm2018}. However, three datasets were excluded due to technical limitations such as missing electrode names and impossibility of downloading the data. The complete list of datasets that were included along with their respective descriptions can be found in \autoref{table:datasets}. The column "No. examples" designates for each dataset the number of examples/trials contained for each combination of subject, session and class. The imagined tasks which form the class label information in each dataset are described using the following abbreviations:
\begin{multicols}{3}
\begin{itemize}
    \item \textit{bh}: both hands
    \item \textit{f}: both feet
    \item \textit{lh}: left hand
    \item \textit{lhrf}: left hand right foot
    \item \textit{n}: navigation
    \item \textit{r}: rest
    \item \textit{ree}: right elbow extension
    \item \textit{ref}: right elbow flexion
    \item \textit{rh}: right hand
    \item \textit{rhc}: right hand close
    \item \textit{rhlf}: right hand left foot
    \item \textit{rho}: right hand open
    \item \textit{rp}: right hand pronation
    \item \textit{rs}: right hand supination
    \item \textit{s}: subtraction
    \item \textit{t}: tongue
    \item \textit{wa}: word association
\end{itemize}
\end{multicols}

It is noteworthy that the datasets included in this study employ diverse imagination strategies. For example, the Lee2019 dataset required subjects to imagine grasping with the appropriate hand. In contrast, the Cho2017 dataset instructed subjects to imagine moving their fingers from the index to the little finger. The Schirrmeister2017 dataset contains executed instead of imagined movements, where subjects engaged in sequential finger-tapping. Of particular note is the Ofner2017 dataset, which includes \textit{single} instances of both executed and imagined movements, diverging from other datasets that sustain or repeat the imagination tasks over several seconds. Some authors did not report the explicit instructions they gave on the imagination strategies and might have left it to the discretion of the subjects.

\subsubsection{Pre-processing}
The pre-processing protocol we used for all the datasets involved setting the trial windows to 0 to 3 seconds post-cue. We strongly limited the EEG channels used to channels C3, C4, and Cz, which are placed over the sensorimotor cortices. The data was resampled at a frequency of 128Hz, aligning with EEGNet's default setting. A bandpass filter with the range 0.5 to 40\,Hz was applied to the data.

\section{Method} \label{method}
\begin{figure}[h]
\centering
\includegraphics[width=\textwidth]{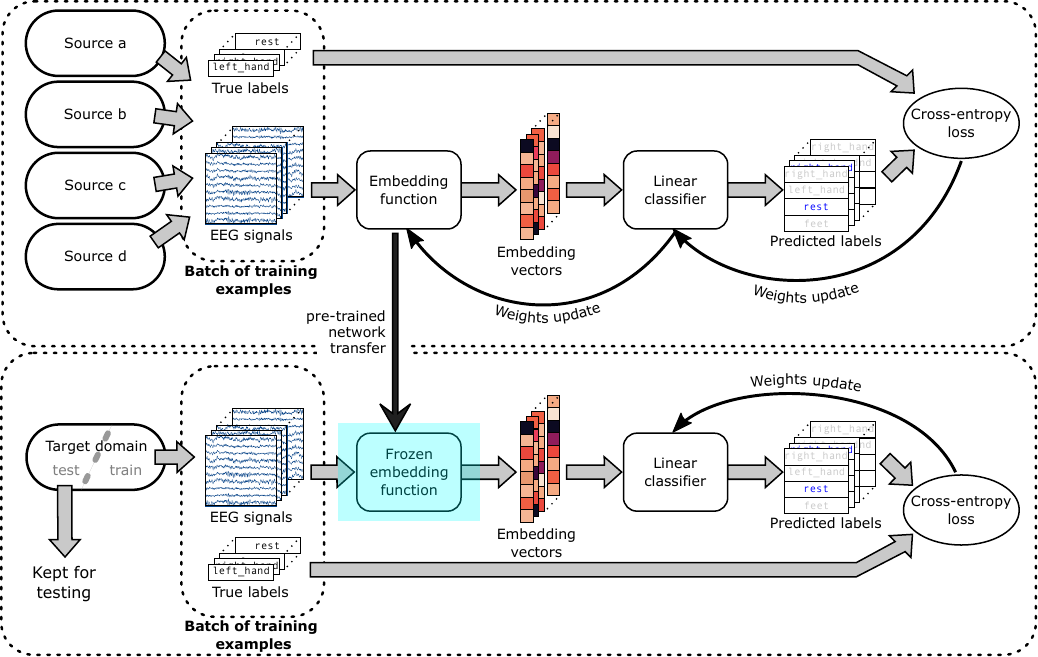}
\caption{Schematic representation of the transfer learning framework we introduced in~\cite{GuePapTan22} and have now extended to cross-dataset transfer. The upper block describes the pre-training phase. The lower block describes the fine-tuning phase.}
\label{fig:diagram}
\end{figure}

In the current study, we employ the transfer learning framework previously proposed in our work~\cite{GuePapTan22}. However, a salient distinction is that, unlike the prior study which focused solely on cross-subject transfer, the present study extends this to cross-dataset transfer. A schematic representation of this methodology is encapsulated in~\autoref{fig:diagram}. This transfer framework has two phases: a pre-training phase using the donor dataset and a fine-tuning phase using the receiver dataset.

\subsubsection{Pre-training Phase}
During the pre-training phase, a neural network architecture is trained from scratch on a set of source datasets, or donors. The aim is to categorize the motor task being imagined in each example/trial. To maintain a simple framework, the hyperparameters chosen are consistent with those recommended by the EEGNet authors for cross-subject training~\cite{lawhernEEGNetCompactConvolutional2018}. Upon completion of this phase, the network's classification head is discarded, and the rest of the architecture is frozen, i.e., the remaining parameters will stay unchanged in the following phase.

\subsubsection{Fine-tuning Phase}
In the subsequent fine-tuning phase, the pre-trained network is utilized as a feature extractor, wherein its parameters are frozen. A following linear classification layer is then trained from scratch to categorize the embeddings produced by the feature extractor. This classification layer is trained and tested using examples from the same session, i.e. the receiver dataset, adopting a within-session cross-validation (CV) approach through multiple random shuffles of the examples. 
Importantly, during the fine-tuning phase, only the linear classification layer is subject to training, while the rest of the architecture remains static. 
The examples used to train the linear classification layer will be called \textit{calibration examples}. The dataset employed for this fine-tuning is interchangeably termed as the \textit{receiver dataset}, \textit{test dataset} or \textit{calibration dataset}, given that both fine-tuning and testing are executed within the same session of a single dataset, even though using separate examples.

\section{Results} \label{results}

\subsection{Experimental Settings}
\subsubsection{Pre-training Phase}
In this phase, a distinct EEGNet architecture was trained for each dataset, utilizing the entire dataset and all available classes therein. Consequently, a total of 12 pre-trained neural network architectures were generated.

\subsubsection{Fine-tuning Phase}
During fine-tuning, we systematically examined varying numbers of calibration examples—specifically 1, 2, 4, 8, 16, 32 (where applicable), and 64 (where applicable) per class—across each session for every subject in the test datasets.
Three distinct analyses were conducted: left hand vs. right hand (\textit{lh-rh}), right hand vs. feet (\textit{rh-f}), and an all-classes classification. The classes specified for each analysis, are the only ones from the receiver used during the fine-tuning phase. The ROC-AUC (area under the receiver operating characteristic curve) metric was adopted for \textit{lh-rh} and \textit{rh-f} classification scores, while accuracy metrics were used for the all-classes scenario. 
The first two binary classification analyses allow for a fair comparison between datasets.
Whereas the classification with all classes is a fallback case in which the chance level varies with the number of classes. We specifically focussed our analysis on the \textit{lh-rh} and \textit{rh-f} class pairs, as they are the most ubiquitously represented across the various datasets.
The number of folds for within-session CV is set to 16. While this may appear excessive, it is crucial to mitigate the significant impact that the selection of just one or two calibration examples can have on the classification performance metrics.

\subsection{Interpretation of Results}
\subsubsection{Data Volume and Analysis}
The three analyses culminated in an extensive set of 871,488 scores, derived from the different CV folds, number of calibration examples, sessions, subjects, test datasets, and pre-trained models.

\subsubsection{Tabulated Scores}
For ease of interpretation, scores were averaged across CV folds, sessions, and subjects. This resulted in three summary tables corresponding to each of the three analytical scenarios. Each table contains an average score for every unique combination of pre-training dataset, test dataset, and number of calibration examples. 
Due to editorial limitations, the main text includes only black-and-white tables restricted to the case with 16 calibration examples per class. 
16 calibration examples per class represents the maximum number that allows for the inclusion of all test datasets in the analysis.
However, the complete color-coded tables are available in the supplementary materials, the location of which was indicated in the introductory \autoref{introduction}.
Results of the \textit{rh-f} analysis with 16 calibration examples per class are in \autoref{table:scores}\subref{table:rh-f}, those related to the \textit{lh-rh} one are in \autoref{table:scores}\subref{table:lh-rh}, and those covering the all-classes one can be located in \autoref{table:scores}\subref{table:all}.

\subsubsection{Graphical Representations}
Line plots showing test scores with respect to the number
of calibration example per class were generated from tabulated results, averaged either across test datasets in \autoref{fig:donors} or pre-training datasets in \autoref{fig:receivers}. Notably, in the averaged line plots, the datasets, even with disparate subject counts, hold equal weight.

\subsubsection{Specific Observations and Cautions}
The crenellated blue line on the right-hand y-axis of \autoref{fig:donors}, denotes the number of test datasets included in each average. Its fluctuation is attributed to the variable number of examples across test datasets which does not allow to include all the datasets when using 32 or 64 calibration examples. Consequently, the average curves' progressions should be interpreted only when the dataset count remains constant.
When the number of test datasets included in the average decreases, an increase or decrease in score can respectively be explained by the disappearance of a "difficult" or an "easy" dataset.

In contrast, when averaging across pre-training datasets, i.e. \autoref{fig:receivers}, the termination of each test dataset line is evident, and in some instances, extrapolation can provide insights into future trends.

It should be noted that scores obtained from the all-classes analyses require cautious interpretation due to varying class numbers across test datasets, which consequently alter the chance levels of accuracy scores.

\newlength{\colwidth}
\setlength{\colwidth}{2ex}

\begin{table}
\caption{Scores using 16 Calibration Examples per Class\label{table:scores}}
\scriptsize
\centering
\subfloat[\textit{f-rh} - AUC Score%
]{%
    \makebox[6cm][c]{\vbox{\begin{tabular}{|l|m{\colwidth}m{\colwidth}m{\colwidth}m{\colwidth}m{\colwidth}m{\colwidth}m{\colwidth}m{\colwidth}|}
\hline
\textbf{Test dataset} & \multicolumn{1}{l}{\rotatebox{90}{AlexMI}} & \multicolumn{1}{l}{\rotatebox{90}{BNCI2014001}} & \multicolumn{1}{l}{\rotatebox{90}{BNCI2015001}} & \multicolumn{1}{l}{\rotatebox{90}{BNCI2015004}} & \multicolumn{1}{l}{\rotatebox{90}{PhysionetMI(I)}} & \multicolumn{1}{l}{\rotatebox{90}{Schirrmeister2017}} & \multicolumn{1}{l}{\rotatebox{90}{Weibo2014}} & \multicolumn{1}{l|}{\rotatebox{90}{Zhou2016}} \\
\textbf{Pretraining dataset} &  &  &  &  &  &  &  &  \\
\hline
AlexMI & 63 & 66 & 57 & 51 & 60 & 81 & 54 & 60 \\
BNCI2014001 & 73 & \bfseries 85 & 71 & 53 & 67 & 92 & 72 & 89 \\
BNCI2014004 & 72 & 70 & 67 & 55 & 63 & 86 & 71 & 85 \\
BNCI2015001 & 70 & 77 & \bfseries 79 & 53 & 67 & 90 & \bfseries 80 & 89 \\
BNCI2015004 & 52 & 65 & 57 & 52 & 56 & 76 & 51 & 60 \\
Cho2017 & 56 & 61 & 57 & 49 & 61 & 70 & 51 & 65 \\
Lee2019\_MI & 74 & 75 & 71 & 51 & 70 & 88 & 78 & 90 \\
Ofner2017(I) & 50 & 50 & 50 & 50 & 50 & 50 & 50 & 50 \\
PhysionetMI(I,E) & \bfseries 78 & 82 & 73 & \bfseries 56 & \bfseries 71 & 92 & 77 & 88 \\
Schirrmeister2017 & 74 & 80 & 75 & 51 & 67 & \bfseries 96 & 77 & 91 \\
Weibo2014 & 71 & 78 & 69 & 54 & 67 & 89 & 72 & 83 \\
Zhou2016 & 71 & 75 & 74 & 53 & 67 & 88 & 78 & \bfseries 95 \\
\hline
\end{tabular}%
}}\label{table:rh-f}%
}\hspace*{.2cm}%
\subfloat[\textit{lh-rh} - AUC Score%
\label{table:lh-rh}%
]{%
   \makebox[3.2cm][c]{\vbox{\begin{tabular}{|m{\colwidth}m{\colwidth}m{\colwidth}m{\colwidth}m{\colwidth}m{\colwidth}m{\colwidth}m{\colwidth}|}%
\hline
\multicolumn{1}{|l}{\rotatebox{90}{BNCI2014001}} & \multicolumn{1}{l}{\rotatebox{90}{BNCI2014004}} & \multicolumn{1}{l}{\rotatebox{90}{Cho2017}} & \multicolumn{1}{l}{\rotatebox{90}{Lee2019\_MI}} & \multicolumn{1}{l}{\rotatebox{90}{PhysionetMI(I)}} & \multicolumn{1}{l}{\rotatebox{90}{Schirrmeister2017}} & \multicolumn{1}{l}{\rotatebox{90}{Weibo2014}} & \multicolumn{1}{l|}{\rotatebox{90}{Zhou2016}} \\
 & & & & & & & \\
\hline
61 & 64 & 62 & 56 & 65 & 64 & 51 & 56 \\
\bfseries 81 & 75 & 66 & 69 & 72 & 65 & 64 & 83 \\
71 & \bfseries 87 & 64 & 70 & 66 & 68 & 69 & 86 \\
72 & 73 & 64 & 66 & 66 & 63 & 64 & 78 \\
58 & 65 & 62 & 55 & 63 & 64 & 50 & 57 \\
66 & 72 & \bfseries 74 & 60 & 70 & 62 & 52 & 65 \\
79 & 82 & 68 & \bfseries 77 & \bfseries 75 & \bfseries 72 & \bfseries 77 & 91 \\
50 & 50 & 55 & 50 & 50 & 50 & 50 & 50 \\
72 & 70 & 65 & 68 & 74 & 65 & 64 & 80 \\
74 & 73 & 64 & 70 & 68 & 69 & 66 & 84 \\
68 & 69 & 65 & 63 & 71 & 64 & 57 & 66 \\
75 & 78 & 66 & 72 & 69 & 70 & 70 & \bfseries 93 \\
\hline
\end{tabular}%
}}%
}\hspace*{.2cm}%
\subfloat[All - Accuracy Sc.%
\label{table:all}%
]{%
   \makebox[3.2cm][c]{\vbox{
\begin{tabular}{|m{\colwidth}m{\colwidth}m{\colwidth}m{\colwidth}m{\colwidth}m{\colwidth}m{\colwidth}m{\colwidth}|}
\hline
\multicolumn{1}{|l}{\rotatebox{90}{AlexMI}} & \multicolumn{1}{l}{\rotatebox{90}{BNCI2014001}} & \multicolumn{1}{l}{\rotatebox{90}{BNCI2015004}} & \multicolumn{1}{l}{\rotatebox{90}{Ofner2017(I)}} & \multicolumn{1}{l}{\rotatebox{90}{PhysionetMI(I)}} & \multicolumn{1}{l}{\rotatebox{90}{Schirrmeister2017}} & \multicolumn{1}{l}{\rotatebox{90}{Weibo2014}} & \multicolumn{1}{l|}{\rotatebox{90}{Zhou2016}} \\
 & & & & & & & \\
\hline
43 & 37 & 23 & 17 & 34 & 46 & 23 & 41 \\
49 & \bfseries 56 & \bfseries 25 & \bfseries 18 & 38 & 58 & 32 & 68 \\
45 & 41 & 23 & 15 & 31 & 53 & 29 & 68 \\
48 & 47 & 23 & \bfseries 18 & 34 & 53 & 30 & 63 \\
34 & 34 & 23 & 16 & 32 & 43 & 22 & 41 \\
36 & 34 & 21 & 16 & 31 & 37 & 20 & 46 \\
48 & 47 & 23 & 17 & 33 & 53 & 32 & 74 \\
33 & 24 & 20 & 17 & 12 & 25 & 14 & 33 \\
\bfseries 56 & 51 & \bfseries 25 & 17 & \bfseries 43 & 60 & \bfseries 36 & 67 \\
54 & 50 & 23 & 17 & 35 & \bfseries 66 & 33 & 69 \\
46 & 46 & 23 & \bfseries 18 & 38 & 54 & 31 & 55 \\
49 & 46 & 24 & 17 & 34 & 54 & 32 & \bfseries 80 \\
\hline
\end{tabular}

}}%
}
\end{table}

\newlength{\figheight}
\setlength{\figheight}{6cm}
\begin{figure}
    \centering
    \begin{tabular}{l}
        \includegraphics[height=\figheight]{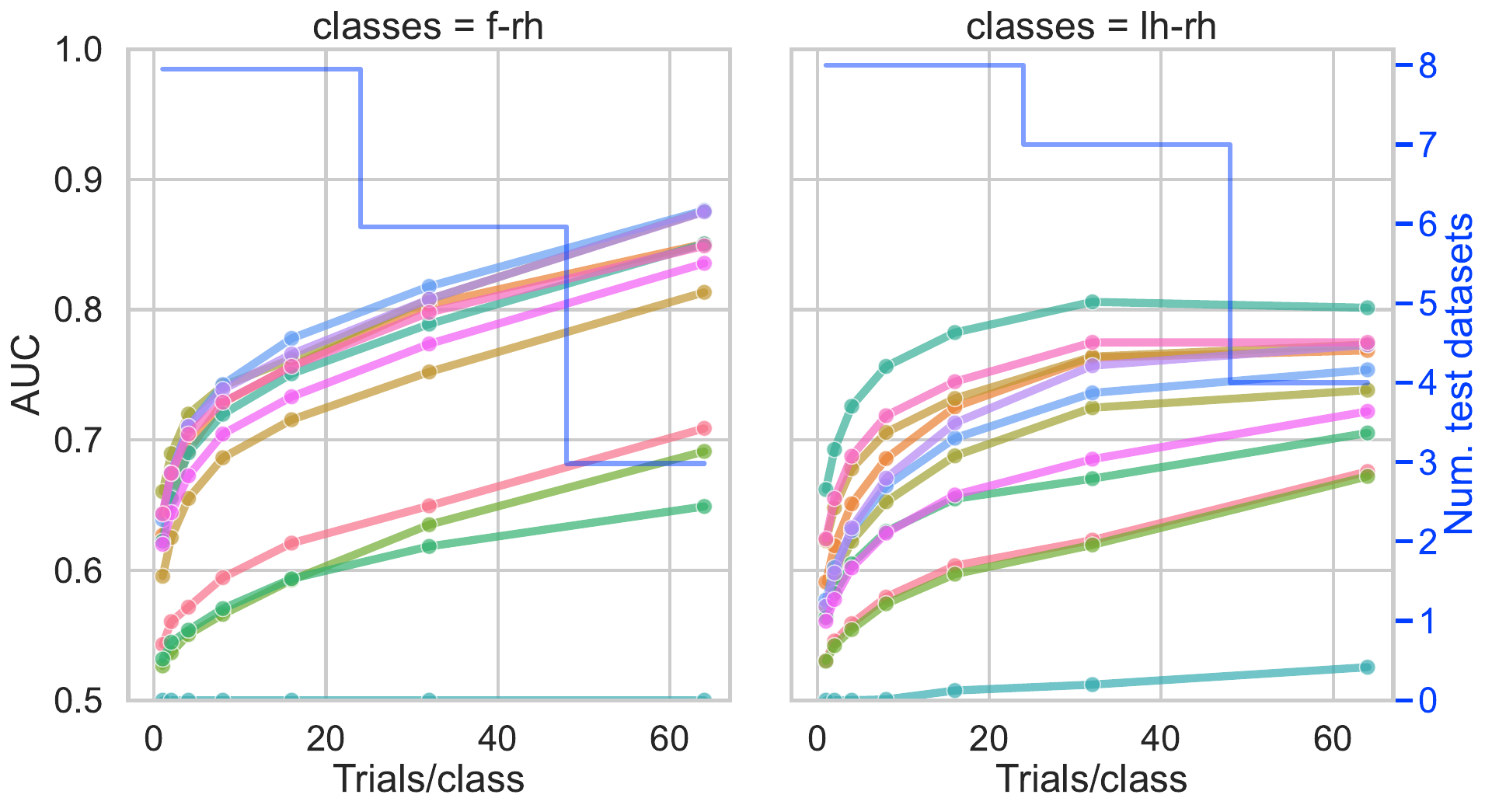} \\
        \includegraphics[height=\figheight]{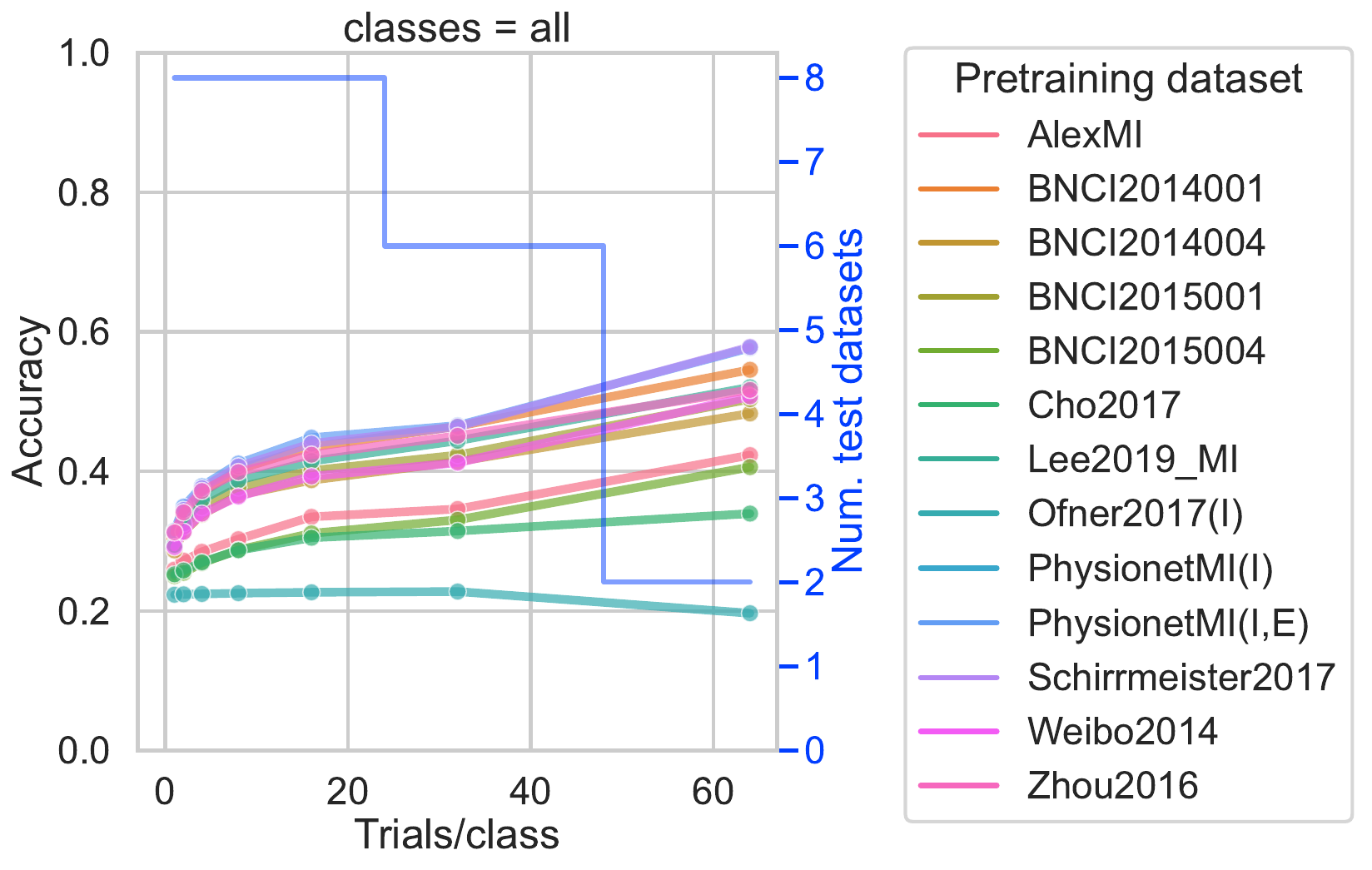}
    \end{tabular}
    \caption{Average Scores across Test Datasets}
    \label{fig:donors}
\end{figure}

\begin{figure}
    \centering
    \begin{tabular}{l}
        \includegraphics[height=\figheight]{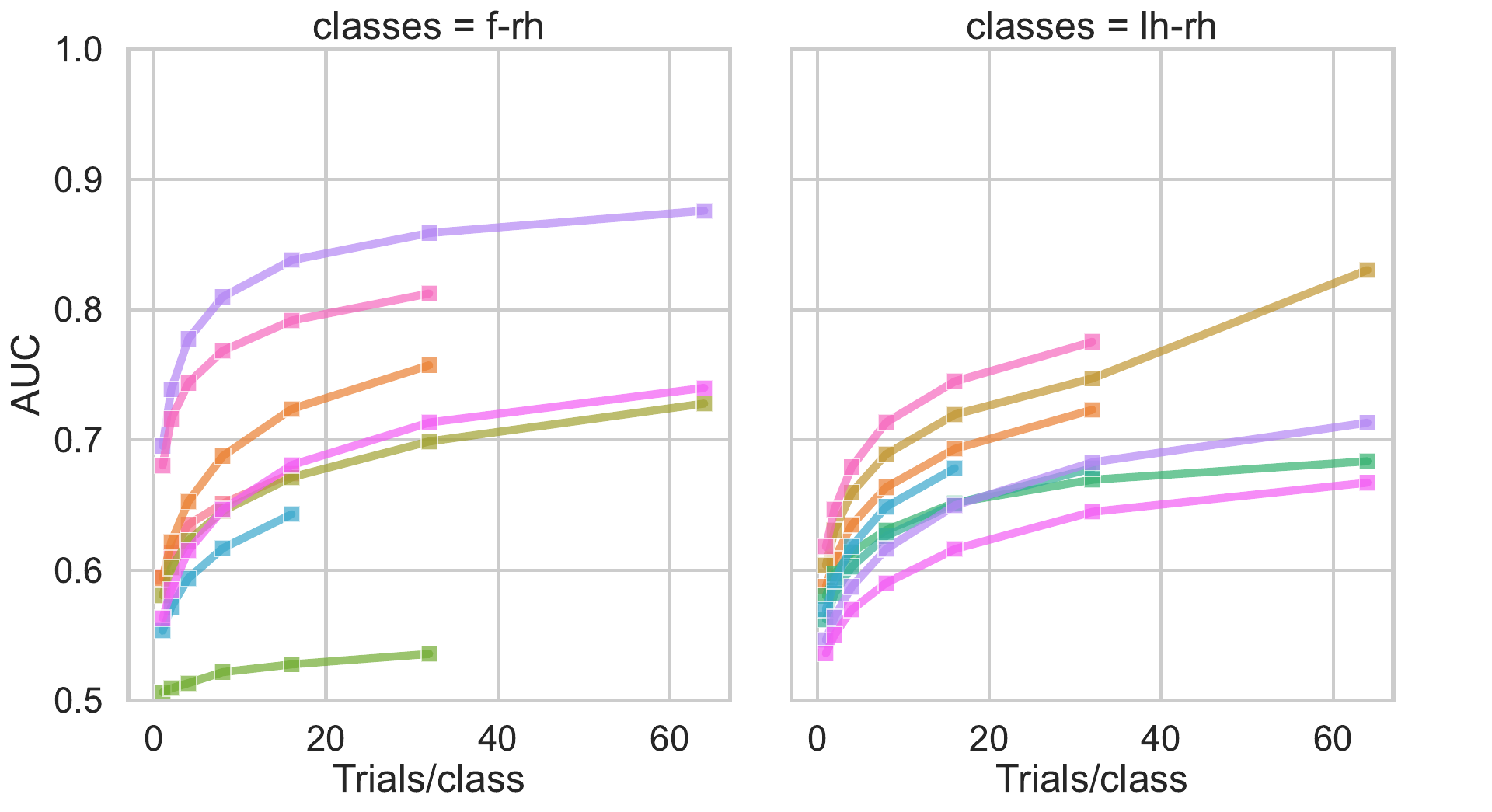}\\
        \includegraphics[height=\figheight]{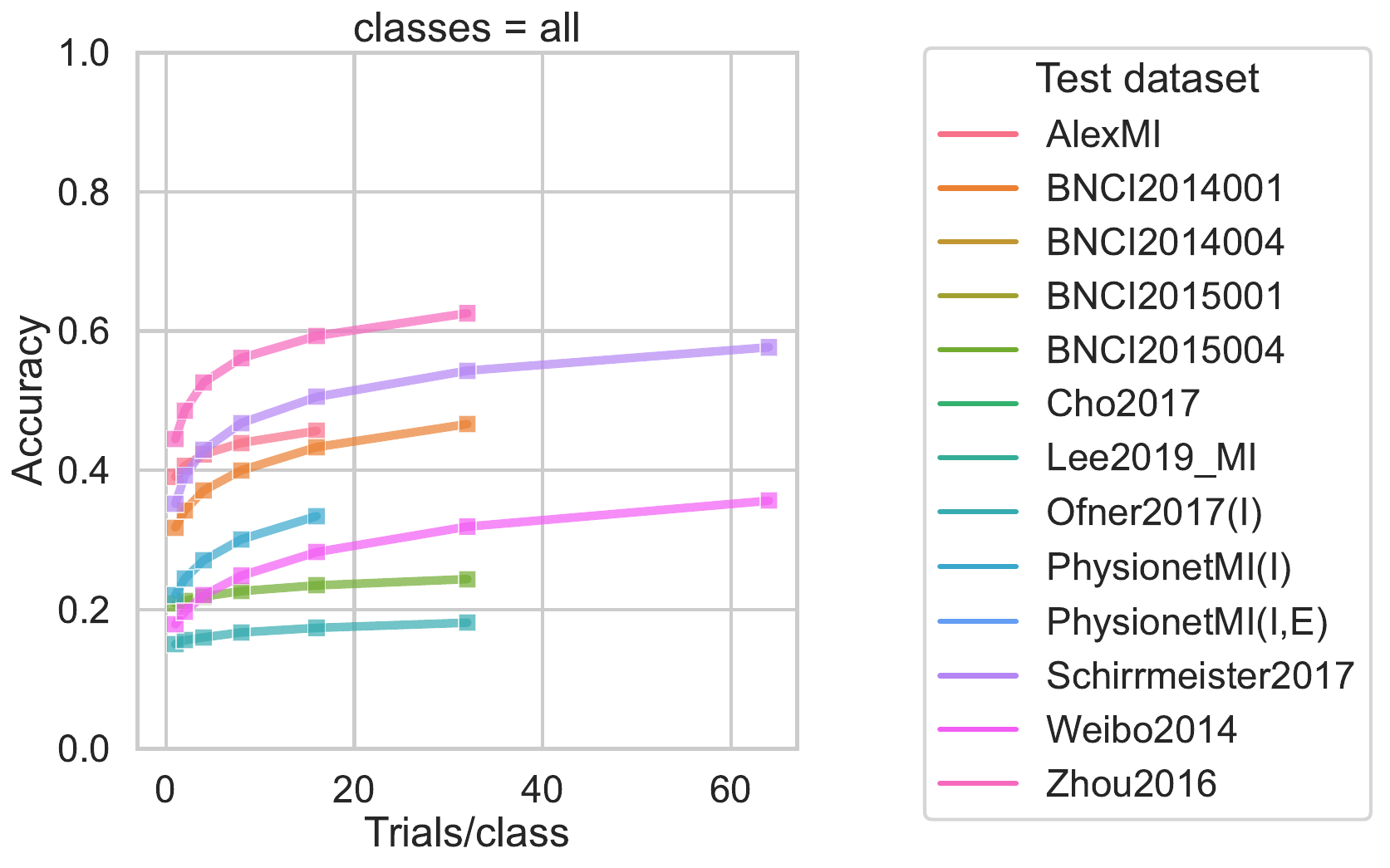}
    \end{tabular}
    \caption{Average Scores across Pre-training Datasets}
    \label{fig:receivers}
\end{figure}

\subsection{Datasets as Donors}
Referring to \autoref{fig:donors}, which illustrates averages across test datasets, the model pre-trained on Ofner2017 consistently performed at chance level. This outcome aligns with the dataset's distinct characteristics, as elaborated in \autoref{datasets}. Across the three classification scenarios, models pre-trained on the donor datasets AlexMI, BNCI2015004, or on Cho2017 underperformed. 
For the \textit{rh-f} classification task, no single model emerged as a clear frontrunner. However, models pre-trained on the donor datasets PhysionetMI, Schirrmeister2017, BNCI2014001, Zhou2016, Lee2019, or BNCI2015001 constituted the leading group.
In the \textit{lh-rh} classification task, the model pre-trained on Lee2019 demonstrated a distinct advantage. 

Scores incorporating all classes from the test datasets are delicate to interpret and are presented primarily for informational purposes.

\subsection{Datasets as Receivers}
Turning our attention to \autoref{fig:receivers}, which displays averages across pre-training datasets, we observe the following:

\paragraph{f-rh analysis}
Schirrmeister2017 emerges as the most straightforward dataset to classify, followed by Zhou2016 and then BNCI2014001. Conversely, BNCI2015004 and PhysionetMI proved most challenging, with performance nearing chance levels for BNCI2015004.

\paragraph{\textit{rh-lh} analysis}
Zhou2016 demonstrated greater ease of classification relative to other datasets, succeeded by BNCI2014004 and then BNCI2014001. The most challenging datasets for this task are Weibo2014, followed by Lee2019, Cho2017, and Schirrmeister2017.

\paragraph{All-classes analysis}
In this context, Zhou2016 and Schirrmeister2017 were the easiest datasets to classify. Notably, none of the pre-trained models supplied features conducive to classifying Ofner2017 and BNCI2015004, with performance for both hovering at chance level.

\section{Discussion} \label{discussion}

\subsection{On the Datasets Evaluated}
Lee2019 emerges as a highly favorable option for pre-training models for a simple \textit{lh-rh} paradigm. Remarkably, even our rudimentary model attains competent classification scores across diverse datasets with very few calibration examples. This is likely attributable to its substantial number of subjects and examples.

PhysionetMI can serve as an interesting benchmark. Its low ranking when averaged across pre-training datasets implies inherent challenges. Moreover, its large number of subjects (109) implies also a large diversity. Furthermore, its positive performance during pre-training suggests that the task encapsulated is not overly divergent from those in other datasets.
For the same reasons, Cho2017 and Lee2019 can also both serve as challenging benchmarks but they are limited to the \textit{lh-rh} paradigm owing to class restrictions. 
Conversely, Zhou2016 and BNCI2014001 appear to be relatively uncomplicated benchmarks, consistently ranking among the top-performing receivers.

Schirrmeister2017 distinguishes itself as both a good donor and receiver. Its performance as a receiver can likely be attributed to the fact that it contains executed motor tasks, which are generally more straightforward to classify than imagined ones. Its strong performance as a donor is probably due to the higher number of examples available per subject, setting it apart from other datasets in the study.

The consistent chance-level performance observed with the Ofner2017 dataset, both as a donor and receiver, aligns with expectations given its marked divergence from the dataset pool, as outlined in \autoref{datasets}. The signal modality reported by its authors more closely resembles an event-related potential (ERP) than the oscillatory desynchronization or synchronization features typically observed for (motor) imagery tasks.

On a final note, pre-training models using a single dataset only probably is sub-optimal, and we opted for this simple approach solely for the purpose of our donor / receiver analysis. Instead, leveraging a diverse collection of donor datasets could enhance model generalizability.

\subsection{On the Transfer Learning Framework}
The transfer learning scheme presented is already widely utilized in the computer vision domain~\cite{dengImageNetLargescaleHierarchical2009}. We posit that its broader adoption within the BCI community would be beneficial for multiple reasons:

First, the proposed framework has deliberately low complexity, eschewing ambiguous choices as we did not converge to it through prior testing. We adopted EEGNet, a common architecture in the BCI landscape, and adhered strictly to author-recommended parameters. Additionally, our reliance on merely three EEG channels reflects the simplicity of the chosen approach.

Our framework successfully accomplishes motor-imagery cross-dataset transfer learning, despite it being known as a  difficult problem. Our results, which are proximal to state-of-the-art algorithms evaluated\footnote{For a comprehensive understanding of the state-of-the-art, the reader is advised to consult the MOABB benchmark~\cite{jayaramMOABBTrustworthyAlgorithm2018}, in conjunction with its forthcoming 2023 update.}
despite using fewer session examples, attest to this. 
Naturally, there is a maximal dissimilarity between datasets after which the transfer becomes impossible. We found that limit with the Ofner2017 dataset.

The efficiency of our framework is emphasized by our expansive analysis involving 871,488 classifiers. Such a broad analysis would be computationally prohibitive without a rapid fine-tuning mechanism.

Following the same idea, the swiftness of our fine-tuning strategy makes it suitable for online scenarios, without necessitating prolonged waiting periods for subject-specific model training as only a linear classification layer needs to be trained, which can be done in a few seconds on any recent CPU.

The modular nature of the two-phased framework allows for independent enhancements, facilitating the seamless incorporation of emerging deep learning techniques in both the embedding and fine-tuning stages.
Because of the current simplicity of all its steps, we can expect large performance enhancement from improvements at any level. 
A first straightforward improvement would be to include additional EEG channels.

Finally, our framework demonstrates a pathway for utilizing deep learning models for BCI decoding. This approach effectively mitigates two challenges inherent to deep learning techniques: the extensive requirement for data and lengthy training durations.

%
%
%
\bibliographystyle{splncs04}
\bibliography{references.bib}

\end{document}